\pdfoutput=1

\documentclass[11pt]{article}

\usepackage{ACL2023}

\usepackage{times}
\usepackage{latexsym}
\usepackage{bm}
\usepackage{url}
\usepackage{hyperref}
\usepackage{microtype}
\usepackage{array}
\usepackage{pifont}
\usepackage{tabularx}
\usepackage{adjustbox}
\usepackage{multirow}
\usepackage{enumitem}
\usepackage{xspace}
\usepackage{tcolorbox}
\usepackage{booktabs,amsfonts,dcolumn,tabulary,tabularx}
\usepackage{amsmath,amsthm,amsfonts,amssymb,bm,stmaryrd,bbm}
\usepackage{makecell}
\usepackage{graphicx}
\usepackage{xcolor,colortbl}
\usepackage{color,soul}
\usepackage{caption}
\usepackage{tikz}
\usepackage{float}
\usepackage{times}
\usepackage{textcomp}
\usepackage{subcaption}
\usepackage{tablefootnote}
\usepackage[T1]{fontenc}

\usepackage[utf8]{inputenc}

\usepackage{microtype}
\usepackage{lipsum}  
\usepackage{inconsolata}

\renewcommand{\paragraph}[1]{\vspace{0.2cm}\noindent\textbf{#1}}

\definecolor{ccon}{HTML}{fee9d4}
\definecolor{cood}{HTML}{d8f0d3}
\definecolor{cid}{HTML}{dae8f5}

\definecolor{gred}{HTML}{cc0200}
\definecolor{ggreen}{HTML}{38761c}

\definecolor{c1}{cmyk}{0,0.6175,0.8848,0.1490}
\definecolor{c2}{cmyk}{0.1127,0.6690,0,0.4431}
\definecolor{c3}{cmyk}{0.3081,0,0.7209,0.3255}
\definecolor{c4}{cmyk}{0.6765,0.2017,0,0.0667}
\definecolor{c5}{cmyk}{0,0.8765,0.7099,0.3647}

\newtcbox{\hlprimarytab}{on line, rounded corners, box align=base, colback=c3!10,colframe=white,size=fbox,arc=3pt, before upper=\strut, top=-2pt, bottom=-4pt, left=-2pt, right=-2pt, boxrule=0pt}
\newtcbox{\hlsecondarytab}{on line, box align=base, colback=red!10,colframe=white,size=fbox,arc=3pt, before upper=\strut, top=-2pt, bottom=-4pt, left=-2pt, right=-2pt, boxrule=0pt}

\newcolumntype{P}[1]{>{\centering\arraybackslash}p{#1}}
\title{Do Large Language Models Know How Much They Know?}

\author{\parbox{\linewidth}{\centering{
    Gabriele Prato$^{1,2,3}$, Jerry Huang$^{1,2,3}$\\ Prasannna Parthasarathi$^{2}$, Shagun Sodhani$^{4}$, Sarath Chandar$^{1,2,5}$} \\
    {\rm $^1$Chandar Research Lab~~$^2$Mila -- Quebec AI Institute~~$^3$Universit\'{e} de Montr\'{e}al\\$^4$Meta FAIR~~$^5$Polytechnique Montr\'{e}al~~$^6$Canada CIFAR AI Chair} \\
    {\rm \texttt{\{gabriele.prato,jerry.huang\}@mila.quebec}}
    }
}

\begin{document}

\maketitle

\begin{abstract}
Large Language Models (LLMs) have emerged as highly capable systems and are increasingly being integrated into various uses. However, the rapid pace of their deployment has outpaced a comprehensive understanding of their internal mechanisms and a delineation of their capabilities and limitations. A desired attribute of an intelligent system is its ability to recognize the scope of its own knowledge. To investigate whether LLMs embody this characteristic, we develop a benchmark designed to challenge these models to enumerate all information they possess on specific topics. This benchmark evaluates whether the models recall excessive, insufficient, or the precise amount of information, thereby indicating their awareness of their own knowledge. Our findings reveal that all tested LLMs, given sufficient scale, demonstrate an understanding of how much they know about specific topics. While different architectures exhibit varying rates of this capability’s emergence, the results suggest that awareness of knowledge may be a generalizable attribute of LLMs. Further research is needed to confirm this potential and fully elucidate the underlying mechanisms.
\end{abstract}

\begin{figure*}[t!]
    \centering
    \includegraphics[width=\textwidth]{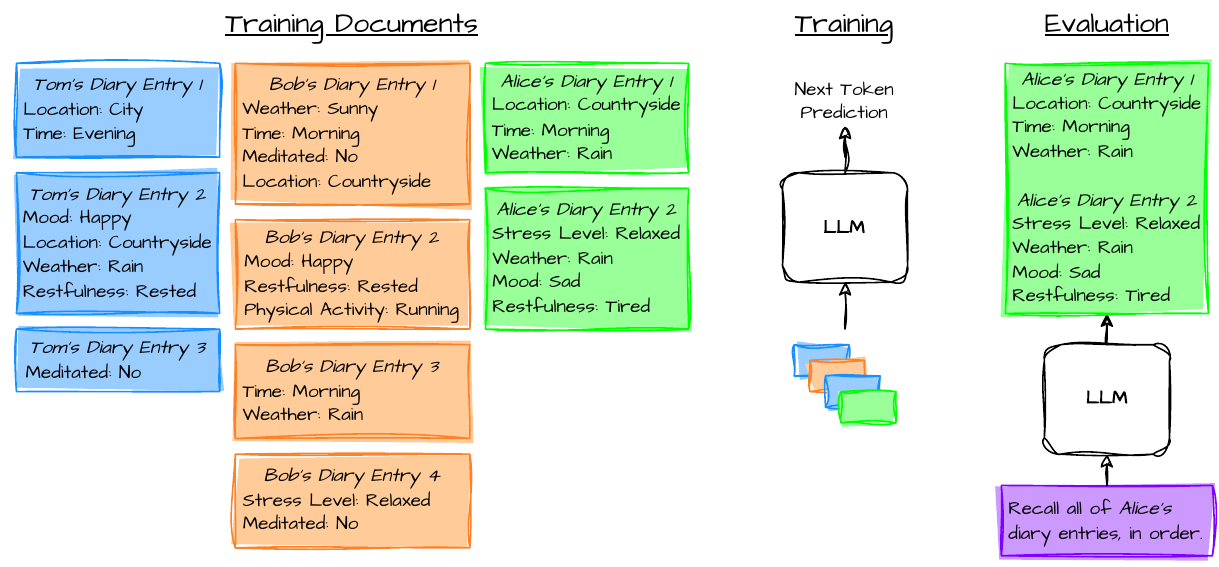}
    \caption{We train LLMs using diary entries from various individuals, with each diarist contributing a random number of entries. We then task the models with recalling all entries written by a specific individual, evaluating their ability to accurately recall the exact number of documents authored by that person.}
    \label{fig:setup}
\end{figure*}

\section{Introduction} \label{sec:intro}
Large Language Models are renowned for their ability to memorize vast amounts of information encountered during training~\cite{GPT4,llama2,gemini}. This information, stored in their parameters, can be recalled during inference, serving both for information retrieval and problem-solving~\cite{2015arXiv150605869V,Radford2019LanguageMA,2022arXiv221011416C,2023arXiv230414767G}. While it is well-established that LLMs can act as knowledge bases~\cite{2019arXiv190901066P,heinzerling-inui-2021-language,2022arXiv220406031A}, the extent to which they understand their own knowledge is less clear~\cite{2024arXiv240115449L}. For instance, do these models know if or when they know the answer to a question~\cite{2022arXiv220705221K,2023arXiv230518153Y}? Can they quantify their own expertise on a topic? Are they aware if some of their knowledge contradicts other information they possess? Can they differentiate between explicitly learned information and implicit knowledge?

These questions are crucial, as awareness of one’s own knowledge and limitations is a vital aspect of any intelligent system. Without it, an AI could be prone to hallucinate~\cite{2023arXiv230906794Y,2024arXiv240111817X}, lie about its expertise~\cite{2023arXiv230413734A,Pacchiardi2023How}, overestimate its responses~\cite{2020arXiv200307892D,GPT4}, or contradict itself~\cite{2023arXiv230708678C}, all of which are undesirable traits for AI systems intended to be useful.

This study focuses on understanding whether LLMs know the extent of their knowledge on specific topics, such as individuals, locations, events or concepts. To explore this, we task LLMs with enumerating everything they know about a given topic—no more, no less. Should a model consistently recall just the right amount of information, it suggests an understanding of the scope of its own knowledge on that topic. Conversely, if a model does not know how much it knows, it may recall too little or hallucinate additional information.

Our approach involves fine-tuning LLMs on the diary entries of various fictitious individuals. Each entry is treated as an individual document in our fine-tuning dataset, with each diarist authoring a random number of entries. During inference, we ask the models to recall all diary entries of a specified individual in chronological order. We then evaluate whether the recalled entries match the original entries both in terms of content and quantity. \autoref{fig:setup} provides an illustrative example.

We benchmark the performance of the OPT~\cite{opt}, Pythia~\cite{pythia}, and Flan-T5~\cite{flan-t5} suites of models. Our key findings are as follows:
\begin{itemize}
    \item All tested LLMs, if scaled sufficiently, demonstrate an understanding of how much they know. This capability appears to emerge at different rates depending on the architecture. For example, an OPT model of a particular size can perform this task effectively if the fine-tuning dataset is sufficiently large, whereas Pythia and Flan-T5 models of the same size require further scaling.
    \item When these conditions are not met (i.e., insufficient scaling), models often recall a random number of diary entries, either recalling too few or hallucinating additional ones.
    \item Interestingly, the number and length of documents do not impact model performance, demonstrating that models are equally effective at memorizing short and long documents, as well as recalling topics associated with a single document or multiple documents.
\end{itemize}
Finally, we discuss potential factors responsible for the observed differences in the emergence of this capability. Overall, our work contributes to a deeper understanding of the inner workings of LLMs, shedding light on a not-so-well-understood aspect of these models.

The insights gained from this research advance our understanding of LLMs, shedding light on their operational capabilities and contributing to the ongoing exploration of their intricate dynamics.

\section{Related Work} \label{sec:related_work}
\subsection{Knowledge Awareness}
Large language models are widely recognized for memorizing a substantial amount of information during their training~\cite{2019arXiv190901066P,roberts-etal-2020-much,10.1162/tacl_a_00324,2022arXiv220207646C,2022arXiv220406031A,2024arXiv240214273H}. However, it remains unclear to what extent these models understand their own knowledge. Research to date has shown that LLMs can assess, with some degree of accuracy, whether they know the answer to a given question~\cite{2022arXiv220705221K,2023arXiv231017918Z,2023arXiv230513712A,2023arXiv230518153Y,2024arXiv240115449L}. 

While these studies primarily evaluate the model's ability to determine if it possesses the knowledge necessary to answer a question, they do not consider the quantity and source of this knowledge. For instance, the question ``Is Jupiter a planet?'' requires knowledge of a single fact, whereas ``Do you know all papers related to topic X?'' necessitates understanding multiple pieces of information, derived from various training samples.

In essence, locating a specific piece of information within a model's parameter space is different from retrieving multiple pieces of information and recognizing when the search is complete. Our research focuses on this latter aspect, seeking to determine whether LLMs comprehend the extent of their knowledge on specific topics.

\subsection{Implicit Knowledge Retrieval}
At the heart of our  methodology lies implicit knowledge retrieval. This involves prompting a model with a question, enabling it to retrieve knowledge stored within its parameters, and subsequently generating an answer based on the retrieved information~\cite{2015arXiv150605869V,2022arXiv221011416C,2023arXiv230414767G}. Considering the black box nature of deep neural networks~\cite{2016arXiv161001644A,2020arXiv201104041S,make3040048,LIANG2021168}, this setup is frequently employed to deduce the inner workings and capabilities of such models~\cite{2021arXiv211208583P,2023arXiv230912288B,2023arXiv230914316A,2023arXiv230914402A,2023arXiv230900667B,2024arXiv240107927M}, offering valuable insights into the knowledge and skills the model has acquired~\cite{2020arXiv200903300H,2021arXiv210703374C,2021arXiv211014168C,2021arXiv210303874H,2023arXiv231112022R}. Hence we consider it to be a fitting analytical approach for our investigation.

\section{Methodology} \label{sec:methodology}
The foundation of our analysis hinges on the ability of models to memorize and recall information. To avoid the influence of existing data, which might be part of the pre-training corpus of the language models we are benchmarking, we generate our own. This ensures that the models have never encountered the data during pre-training, thereby preventing any contamination of our results.

In essence, our approach involves: (i) {\bf generating} the training documents, (ii) {\bf fine-tuning} a language model using its pre-training objective to memorize these documents, and (iii) {\bf testing} the language model's ability to recall all related documents. We delineate each stage of our framework in the following sections.

\subsection{Data Generation}
Given $N$ diarists, where $N$ is a hyperparameter, we generate a random number of diary entries for each diarist, following the template:
\begin{quote}
    \texttt{\hspace{-2pt}\{name\}'s Diary Entry \{i\}}\\
    \texttt{\{attribute\}}\\
    \texttt{\vdots} \\
    \texttt{\{attribute\}}
\end{quote}
where \texttt{\{name\}} is the diarist (e.g., ``Tom'') and \texttt{\{i\}} is the entry number (e.g., ``1''). The document contains a random number of \texttt{\{attribute\}}, each selected randomly without replacement from a set (\autoref{tab:attributes}), along with a randomly chosen value (e.g., ``Time: Morning''). Additionally, for each individual, we have one question following the format:
\begin{quote}
\texttt{Recall all of \{name\}'s diary entries, in order.}
\end{quote}
The answer to the question is the concatenation of the individual's diary entries in ascending order by entry numbers. \autoref{fig:setup} illustrates examples of both generated documents and questions.

To effectively train the model, we incorporate 90\% of the question-answer (Q/A) pairs, along with \textit{all} diary entries, into the training set. The remaining 10\% of the questions are evenly divided into a validation set and a test set. By adding Q/A examples to the training set, the model can learn the evaluation task, similar to the process of instruction-tuning.

Initially, we trained the model first on the documents and then on the evaluation task. However, this approach led to catastrophic forgetting of the documents and overfitting on the Q/A examples. Therefore, we decided to fine-tune the model on both simultaneously to prevent these issues.

\subsection{Fine-Tuning \& Evaluation}
To benchmark an LLM, we begin by fine-tuning it using its pre-training objective, such as causal language modeling, on our training set. This fine-tuning process mirrors the standard training of an LLM on a text corpus. Depending on the architecture of the LLM, we format the input as follows:
\begin{itemize}
    \item \textbf{Decoder-Only Models (e.g., OPT):} For both diary entries and Q/A pairs, the training objective is causal language modeling. In the case of Q/A pairs, we concatenate the question with the answer into a single text sequence, separated by an end-of-line token (`\textbackslash n').
    \item \textbf{Encoder-Decoder Models (e.g., Flan-T5):} When processing a diary entry, the first line (e.g., ``Tom's Diary Entry 1'') is input to the encoder, and the decoder generates the \textit{entire} document. For Q/A pairs, the question is fed to the encoder, and the decoder predicts the answer.
\end{itemize}
Throughout the fine-tuning process, we periodically evaluate the model on the validation set. For decoder-only architectures, the model is prompted with a question with the goal of generating the corresponding answer. For encoder-decoder architectures, the question is given to the encoder and the decoder must produce the answer.

We fine-tune up until the validation performance plateaus. We then select the best checkpoint based on peak validation performance, and evaluate the model on our test set using the same procedure as with the validation set. Performance is measured in terms of accuracy, defined as the number of correctly answered questions. An answer is deemed correct if it matches the ground truth exactly, with no errors in the number of documents recalled and the content of each recalled document.

\subsection{Design Motivation}
Requiring the model to consolidate information from multiple training documents allows us to assess whether it understands the extent of its knowledge related to the individual in question. Specifically, during training, the model memorizes the diary entries. Then, in the evaluation phase, it needs to know how many documents to recall, meaning the model must know how many diary entries it knows about the individual. If a model consistently recalls the exact number of documents, it demonstrates an understanding of the scope of its knowledge regarding that individual. Conversely, a model which does not know how many documents it knows, would recall a random number.

As for our choice of using synthetic data, it allows us to precisely control its distribution and properties. This extends to the length and content of the documents, as well as the number of diary entries authored by an individual. By using attributes as the body of the documents, we can manage the entropy, ensuring that each sentence contains a fixed amount of information. Consequently, adding an additional sentence consistently increases the document's information by that fixed amount.

This approach enables us to examine how document length affects the model in a more controlled manner compared to using real data. While we have arbitrarily chosen individuals as the topic linking multiple documents, this could have been any other concept. We believe this choice does not impact the observed trends in the results.

Overall, our benchmark is designed to facilitate the study of this problem and its key variables in a controlled environment, emulating the challenge faced by language models of memorizing information during training and understanding the extent of their knowledge concerning specific topics.

\section{Experiments} \label{sec:experiments}

\begin{figure*}[t]
    \centering
    \includegraphics[width=\textwidth]{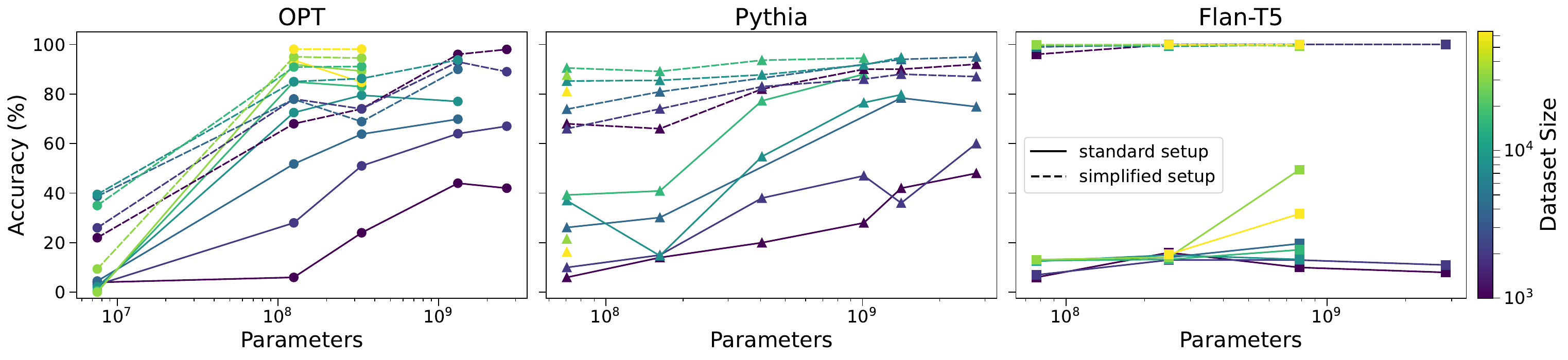}
    \caption{Accuracy of various models on our benchmark is depicted with \textit{solid} lines, where each line represents a different model suite (e.g., OPT) ranging from the smallest to the largest variant, fine-tuned on datasets of varying sizes as indicated by the line colors. For comparison, models trained under a simpler setup are shown with \textit{dashed} lines, where all information necessary to answer a question is contained within a single training document, eliminating the need to recall information from multiple documents.}
    \label{fig:unseg_vs_seg}
\end{figure*}

\begin{figure*}[t]
    \centering
    \includegraphics[width=\textwidth]{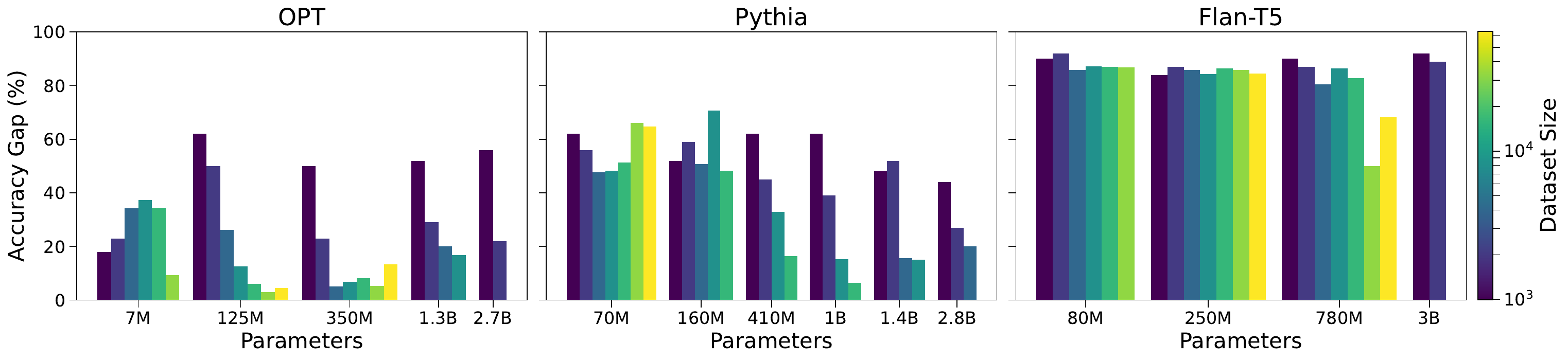}
    \caption{
        Gap in accuracy between the standard and simplified setup in \autoref{fig:unseg_vs_seg}, for a same sized model trained on a same sized dataset.The effect of scaling the dataset and model size varies greatly depending on the architecture.
    }
    \label{fig:unseg_vs_seg_gap}
\end{figure*}

\subsection{Setup} \label{sec:setup}
\paragraph{Dataset.} To evaluate the impact of the number of training examples on the model performance, we generate six datasets containing 1K to 64K diarists, with each successive dataset doubling in size compared to its predecessor. By incrementally enlarging the dataset size as described, models see a broader array of examples from which they can learn to derive their generative capabilities, while simultaneously being challenged to memorize a larger volume of documents.

For each individual, we generate 1 to 8 diary entries, with each entry consisting of 1 to 8 attributes. The training, validation and test sets each contain an equal distribution of individuals who have written one, two, three, etc. diary entries. Similarly, we maintain a uniform distribution for document lengths. Dataset details, such as the number of authors, diary entries, and Q/A pairs, are provided in \autoref{app:dataset_details}.

\paragraph{Models.} We benchmark the following suit of publicly available models: decoder-only OPT (7M to 2.7B)~\citep{opt} and Pythia (70M to 2.8B)~\citep{pythia}, and encoder-decoder Flan-T5 (80M to 3B)~\citep{flan-t5}. A comparison of these architectures is provided in \autoref{app:model_dif}. Training hyper-parameters are provided in \autoref{app:training_details}. Unless specified otherwise, reported metrics are based on the test set.

\subsection{Results}

\paragraph{Effect of Architecture \& Scale.} We first evaluate the impact of architecture, model size, and dataset size on performance. We fine-tune each model on our datasets and report their performance as solid lines, labeled as `standard setup' in \autoref{fig:unseg_vs_seg}. The horizontal axis represents model size, the vertical axis indicates the percentage of correctly answered questions, and the line color signifies the dataset size. Each line on the plot corresponds to a specific architecture (e.g., OPT), ranging from the smallest to the largest model, trained on a particular dataset size. Due to the significant computational cost associated with these experiments, we do not exhaustively explore all combinations of dataset and model size. Instead, we focus on a representative subset of combinations, sufficient to analyze and discern the key trends effectively.

For the OPT suite, we observe a general trend where performance improves as both model size and dataset size increase. Beginning with the smallest variant, which consists of 7M parameters, performance initially improves as the dataset expands, peaking at 4K diarists. However, beyond this threshold, further scaling of the dataset leads to a decline in performance. This pattern suggests that while larger datasets enhance generalization, there comes a point where the model's capacity becomes saturated, causing diminishing returns or even a drop in effectiveness.

In contrast, the 125M-parameter OPT model demonstrates a markedly different behavior. This model is sufficiently large that increasing the dataset size up to the maximum tested---64K diarists---results in consistent performance improvements. The difference in performance between this model trained on the smallest versus the largest dataset is particularly striking, highlighting the significant impact of dataset scaling when paired with a model of sufficient capacity.

Furthermore, increasing the model size while keeping the dataset size constant generally leads to performance gains.

The Pythia models exhibit a similar trend to the OPT suite, where performance improves as both model size and dataset size increase. However, an interesting distinction emerges when comparing the two architectures: performance gains appear sooner in OPT models than in Pythia. Specifically, the 125M-parameter OPT model significantly outperforms the 160M-parameter Pythia model when trained on our largest datasets. This discrepancy underscores differences in how quickly the studied capability emerges depending on the underlying model architecture.

Finally, the performance of Flan-T5 models exhibits a distinct pattern compared to the other architectures. On the smallest datasets, increasing model size alone does not lead to any noticeable improvements. Performance gains only begin to emerge at 783M parameters, and even then, only when trained on the two largest datasets.

Due to computational limitations, we were unable to test the largest Flan-T5 model, with 2.8B parameters, on our largest datasets. However, the overall results suggest that this capability does indeed emerge with sufficient scale---though the rate at which it develops varies depending on the model architecture.

\paragraph{Effect of Distributed Information.} We compare the model performance against a second set of models trained in a simpler setup. Particularly, this second group of models is trained on identical datasets, but with all diary entries authored by the same individual merged into a single training document rather than each entry being its own document. This approach is equivalent to training the models on the answers directly, requiring them to simply memorize and recall single documents. The performance gap between these two setups highlights the added difficulty of dealing with information spread across multiple training documents. This distribution could affect how information is stored in the model's parameters, potentially making it harder for the model to consolidate it when it is dispersed.

In \autoref{fig:unseg_vs_seg}, the results of training within this more straightforward setup are shown as dashed lines, labeled `simplified setup'. In all cases, these models exhibit significantly improved performance compared to the same base model trained within the distributed setup. Interestingly, all Flan-T5 models achieve near-perfect accuracy in this simplified setup whereas OPT and Pythia suites do not, despite performing well and improving with scale.

To better illustrate the performance gap between both setups, we provide a clear visualization in \autoref{fig:unseg_vs_seg_gap}. The vertical axis shows the accuracy gap between the `simplified' and `standard' setup, for models of the same size, trained on datasets containing the same number of individuals. Results are grouped by model size, with colors denoting dataset size.

For the OPT models, the gap narrows as the dataset size increases, with the exception of the smallest model. In the case of Pythia, the gap only seems to narrow for larger models trained on sufficiently large datasets. Lastly, for Flan-T5, the performance gap barely shrinks as both dataset and model size scale, with the exception of the 780M parameter model trained on the largest datasets.

It remains unclear why Flan-T5 models perform so well in the simpler setup but so poorly in the standard setup. Given that the model has near perfect accuracy in the prior, its poor performance in the latter cannot be attributed to an issue in the methodology, as the process is the same in both cases. The only difference is that, in the latter case, the model must recall information from multiple documents rather than a single one. Therefore, the model specifically has an issue with this aspect.

For all models, it is uncertain whether their performance in both setups will continue to improve with scale and if the gap will eventually disappear.


\begin{figure*}[t!]
    \centering
    \includegraphics[width=\textwidth]{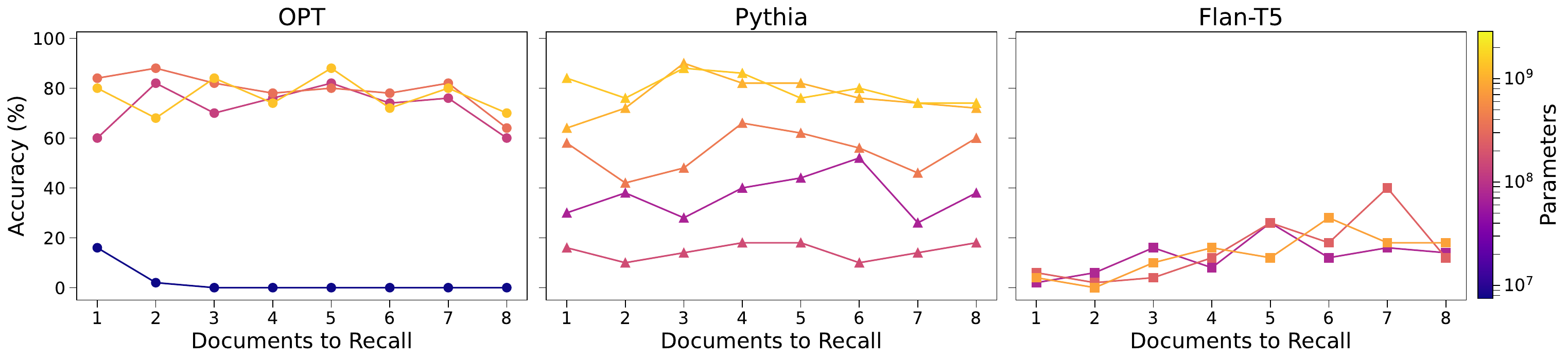}
    \caption{Impact of the number of documents needing to be recalled on the likelihood of a model's answer containing an error. Results are from models trained on our 8K dataset. Surprisingly, we observe no significant difference in performance as the number of documents to recall increases.}
    \label{fig:num_docs_x_accuracy}
\end{figure*}



\begin{figure*}[t!]
    \centering
    \includegraphics[width=\textwidth]{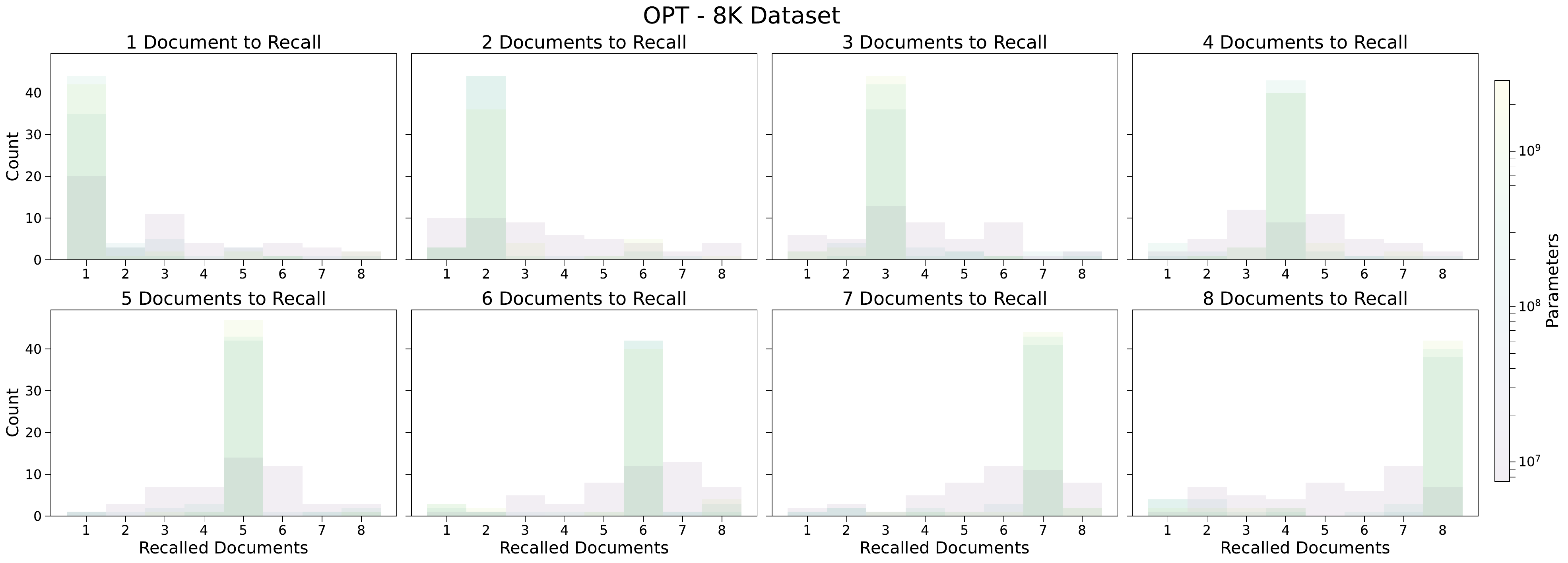}\\[10pt]
    \includegraphics[width=\textwidth]{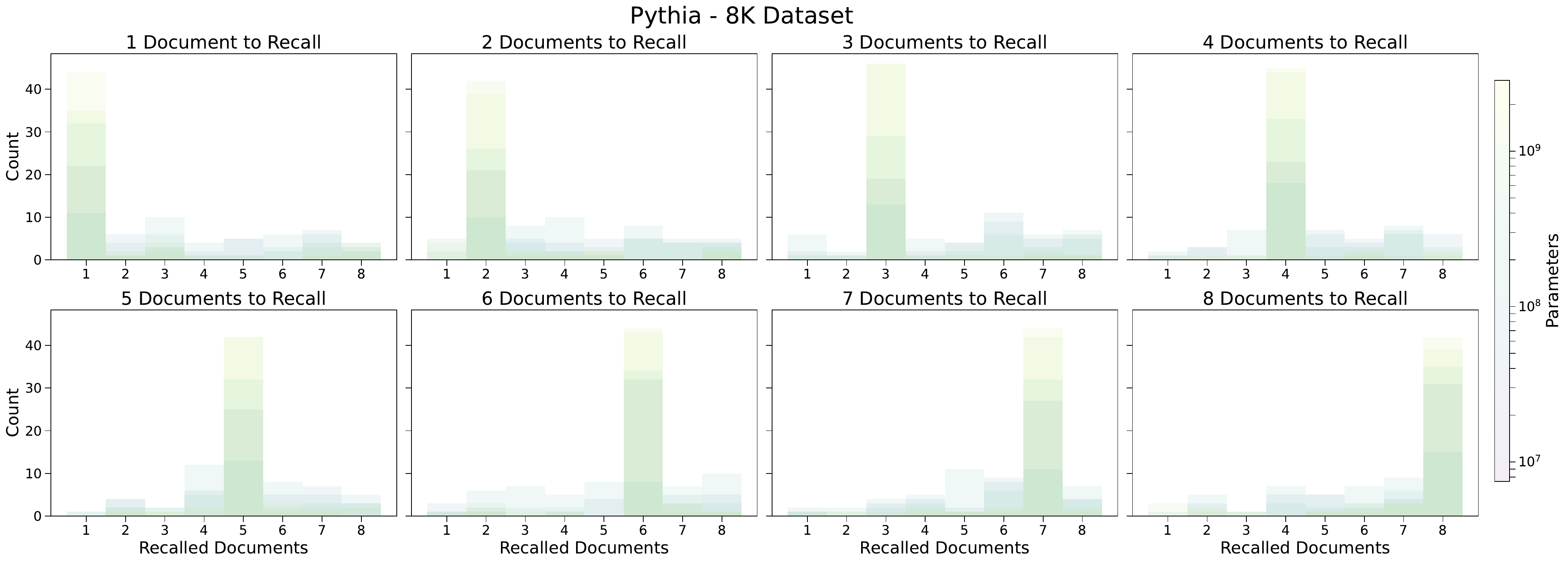}
    \caption{Number of documents recalled by each model in comparison with the target. Color indicates model size. Results are on our 8K dataset. We observe that as the model scales, its ability to recall the correct number of documents emerges.}
    \label{fig:hist_num_docs}
\end{figure*}

\begin{figure*}[t!]
    \centering
    \includegraphics[width=\textwidth]{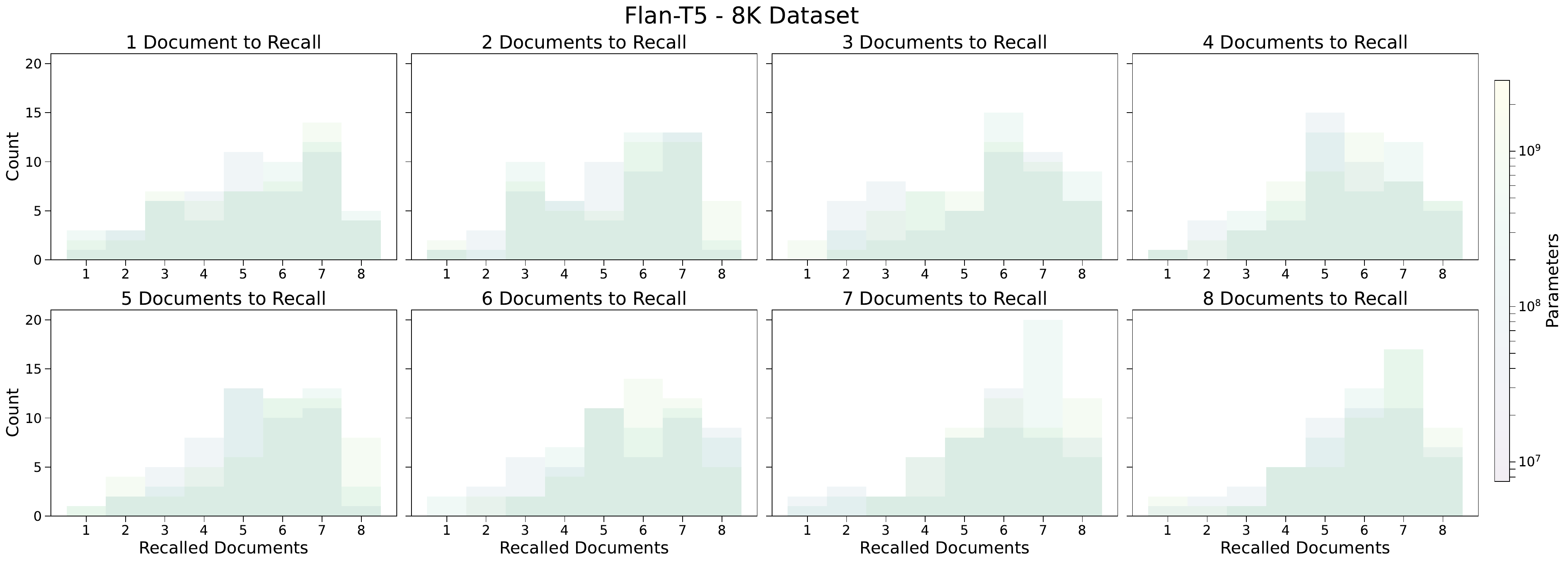}\\[10pt]
    \includegraphics[width=\textwidth]{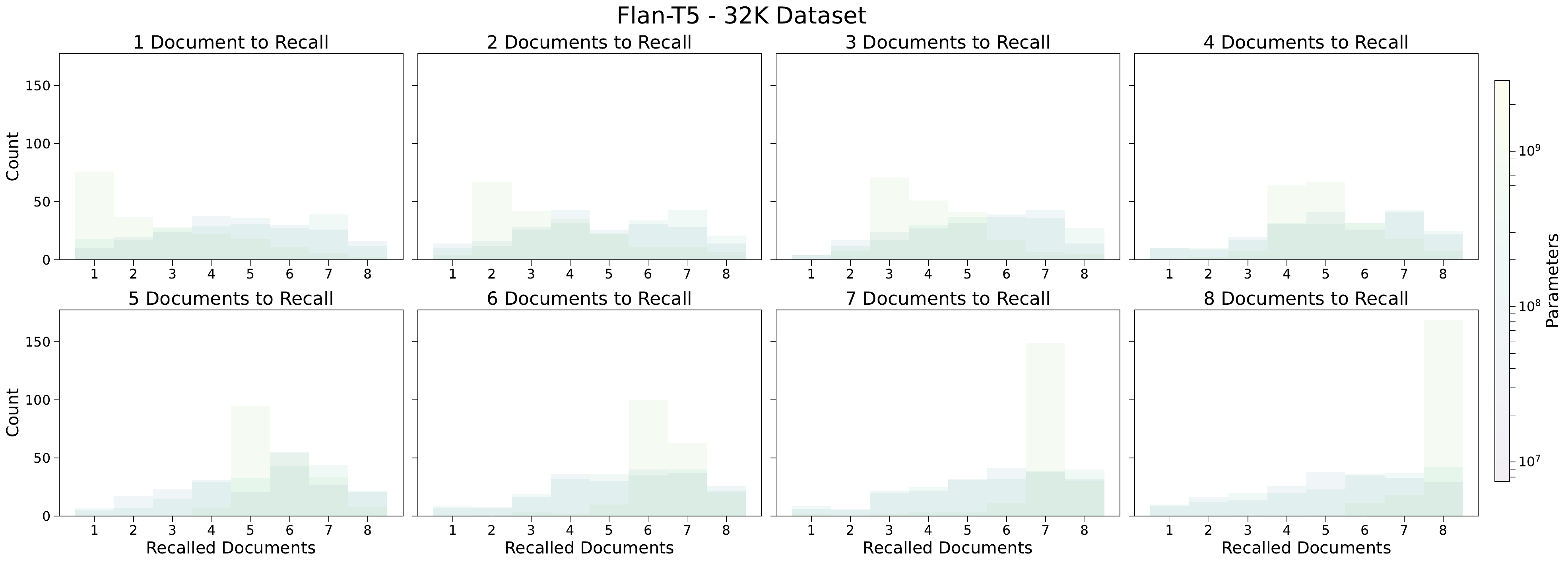}
    \caption{Number of documents recalled by Flan-T5 models compared to the target. Color indicates model size. Results are shown for both the 8K and 32K datasets. On the 8K dataset, recall appears inconsistent regardless of model size, whereas on the 32K dataset, recall improves with scaling. This highlights the importance of increasing both model and dataset size.}
    \label{fig:hist_num_docs_flan}
\end{figure*}

\paragraph{Effect of Number of Documents.} Next, we explore how the number of documents to be consolidated and recalled impacts model performance. In \autoref{fig:num_docs_x_accuracy}, we report accuracy grouped by the number of documents in the target answer (horizontal axis). Line color indicates model size. To maintain clarity, we display only the performance of models trained on the 8K diarist dataset, as the observed trends are consistent across other datasets. Notably, there are no results on the simpler setup in this and further analyses.

Surprisingly, models do not demonstrate a decline in performance when more diary entries need to be recalled. Given the increased content to be generated, one might expect a higher propensity for errors in the model answers. However, this observation could be attributed to the model's capacity being sufficient, and performance deterioration might only appear when recalling a much greater number of documents.

To gain deeper insights into model behavior, we analyze the number of documents recalled by the models in comparison with the target number of documents (\autoref{fig:hist_num_docs} \& \ref{fig:hist_num_docs_flan}). 

For both OPT and Pythia models trained on a dataset of 8K diarists, smaller models appear to recall a random number of documents. However, as model size increases, the ability to accurately determine the appropriate number of documents to recall emerges.

In contrast, Flan-T5 models trained on the same 8K-diarist dataset consistently retrieve a seemingly random number of documents, regardless of model scale. Interestingly, when scaling up to a dataset of 32K diarists, Flan-T5 exhibits a pattern similar to that of OPT and Pythia---where the capability to recognize how many documents should be recalled emerges as model size increases.


\begin{figure*}[t!]
    \centering
    \includegraphics[width=\linewidth]{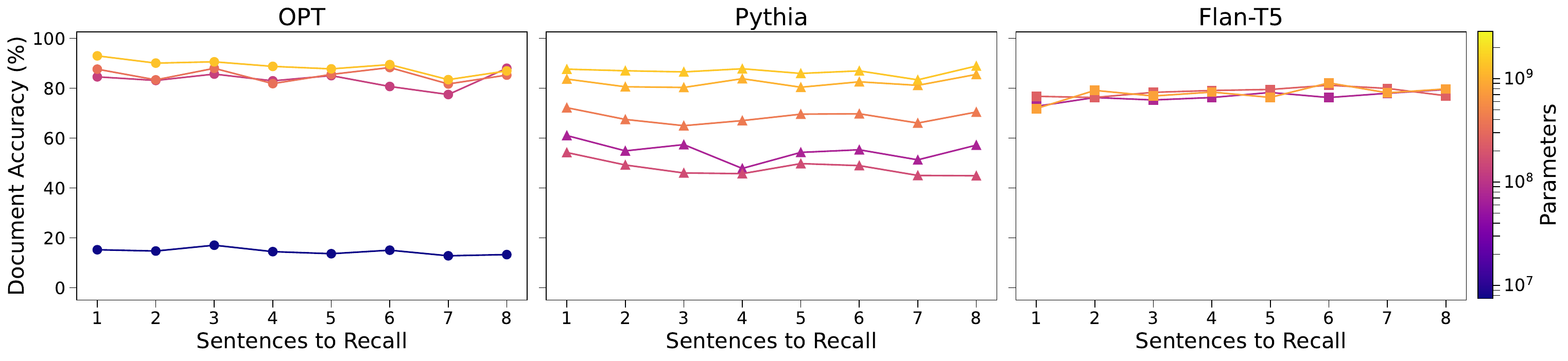}
    \caption{Taking the subset of documents in a model's answer, which are also in the target, we measure the number of such documents that are free of errors, defined as the `document accuracy'. We then categorize this rate by the length of the corresponding target document, in order to measure its effect on the recall capabilities of the model. Results are from models trained on the 8K dataset. Peculiarly, we find that documents recalled by models aren't more likely to contain errors as their length increases.}
    \label{fig:doc_len_x_accuracy}
\end{figure*}


\begin{figure*}[t!]
    \centering
    \includegraphics[width=\textwidth]{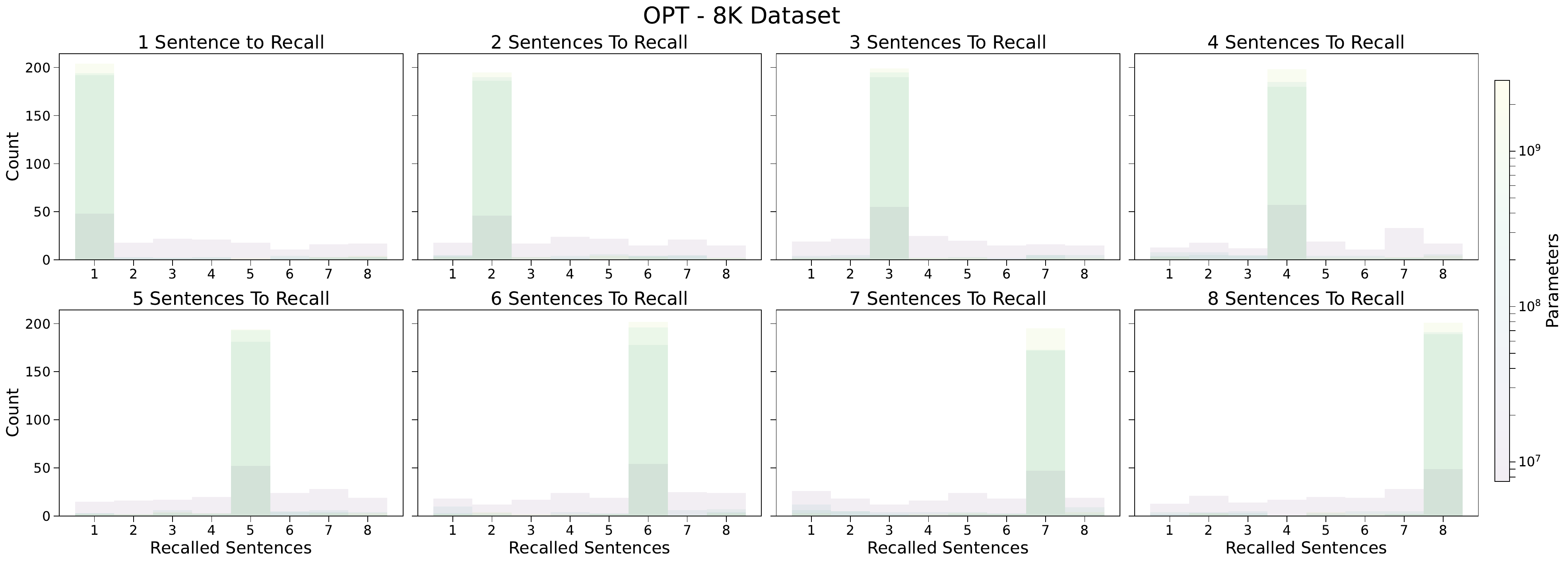}\\[10pt]
    \includegraphics[width=\textwidth]{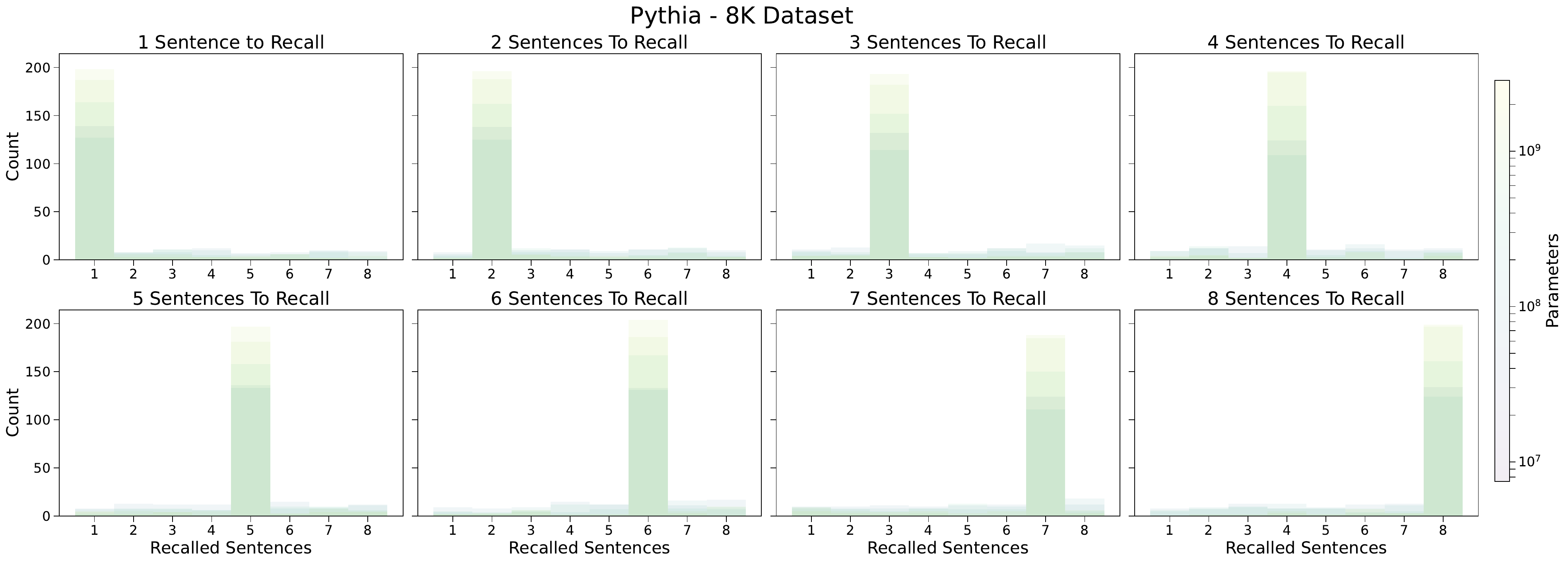}\\[10pt]
    \includegraphics[width=\textwidth]{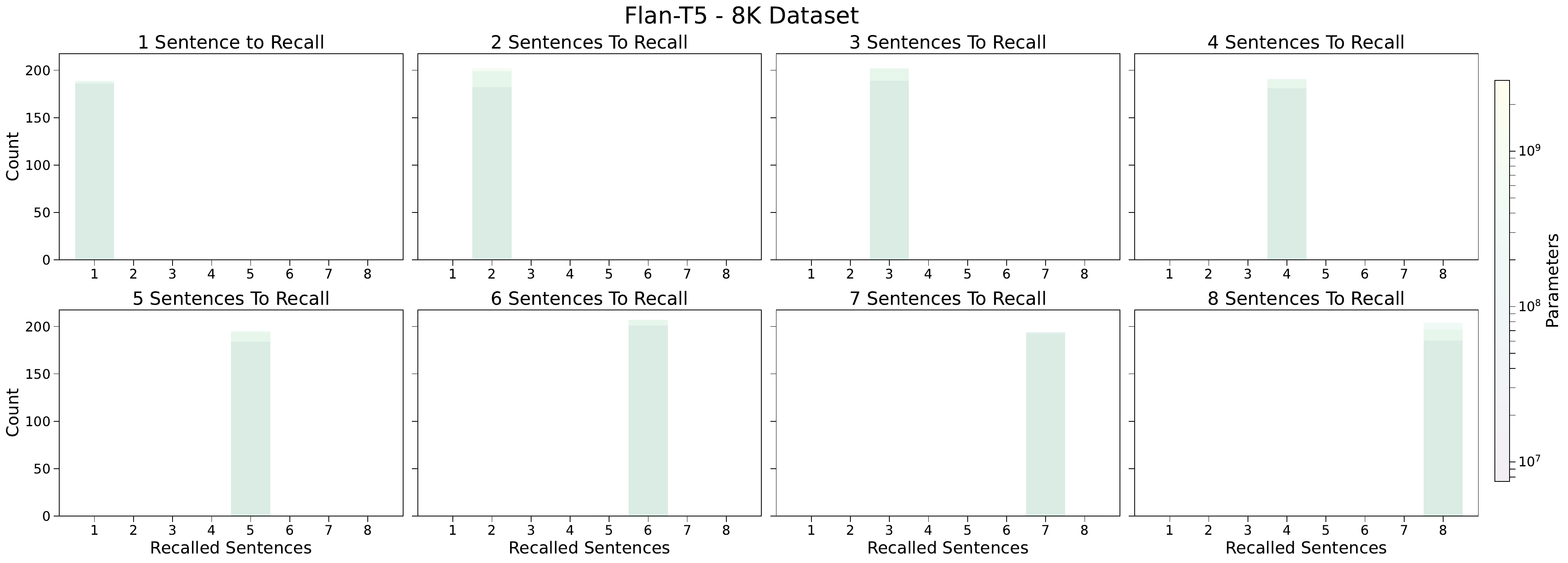}
    \caption{Number of sentences to be recalled in a document compared to the target number of sentences. Color indicates model size. Results are from models trained on our dataset with 8K diarists. Models are able to properly recall the correct number of sentences given sufficient scale.}
    \label{fig:hist_num_sen}
\end{figure*}

\paragraph{Effect of Document Length.} Previously, our method for measuring accuracy involved counting the number of model answers that matched the target answer exactly. We now shift our focus to evaluating the accuracy of individual documents within a model's answer, which we refer to as \textit{document accuracy}.

In this analysis, we only consider the documents recalled by the model that are also present in the target answer, regardless of whether these documents are correct. Our objective is to examine how the length of the target documents influences the model's ability to recall them accurately. Hence, we restrict our analysis to this specific subset of documents, as we need a target for their length.

For these selected documents, we count those that are free of errors and represent this rate on the vertical axis of \autoref{fig:doc_len_x_accuracy}. The performance is categorized by the target length on the vertical axis, and the line color indicates the size of the model. Once again, to maintain clarity, we display only the performance of models trained on the 8K diarist dataset.

Across all models, performance appears unaffected by document length, despite the expectation that longer documents might introduce more errors. One possibility is that the models are sufficiently large to handle even the longest tested documents without performance degradation. However, we hypothesize that much longer documents may eventually impair performance, which would require further testing.

To further understand model behavior, we analyzed the number of recalled sentences, in comparison with the target document length. The histograms in \autoref{fig:hist_num_sen} illustrate these distributions for each model, with color indicating the model size. Similar to our analysis of the number of recalled documents, we observe that model performance improves with scale: smaller models appear to recall a random number of sentences, whereas larger models consistently recall the exact number. Interestingly, however, the smallest Flan-T5 models recall the correct number of sentences, suggesting that they are better at memorizing documents than OPT and Pythia models of the same size.

\paragraph{Investigating Performance Discrepancies.} Our results indicate that the ability to consolidate and accurately recall the correct number of documents varies depending on the model suite, but the underlying reasons for this discrepancy remain unclear. At a high level, these differences in performance could be due to several factors: architectural variations, the effectiveness of pre-trained weights for fine-tuning on this task, the fine-tuning hyperparameters, or a combination of these elements.

To investigate this further, we fine-tuned an OPT-125M, a Pythia-70M, and a Flan-T5 Small model, all with randomly initialized weights, using our dataset with 32K diarists. We then compare their performance against the pre-trained models that were fine-tuned on the dataset of the same size.

Our findings reveal that the Pythia model initialized with random weights significantly outperform the pre-trained weights (\autoref{tab:scratch_vs_pretrained}). This suggests that architectural differences are not responsible for the poor performance of this model. Instead, the issue lies in the capability of the pre-trained weights to be effectively fine-tuned for this specific task.

\begin{table}[t!]
  \centering
  \begin{tabular}{lcc}
    \toprule
    Model & Pre-trained & Scratch \\
    \midrule
    OPT-125M & \textbf{91.94\%}  & 82.19\% \\
    Pythia-70M & 21.56\% & \textbf{45.50\%} \\
    Flan-T5 Small & \textbf{13.12\%} & 12.06\% \\
    \bottomrule
  \end{tabular}
  \caption{Comparison between fine-tuning pre-trained models versus training models initialized with random weights on the 32K dataset.}
  \label{tab:scratch_vs_pretrained}
\end{table}

Regarding Flan-T5, fine-tuning with randomly initialized weights does not appear to enhance performance when compared to fine-tuning the pre-trained model. This observation suggests that the model's architecture is responsible for the observed differences in performance.

Although fine-tuning hyperparameters could also be a factor, we conducted a thorough search. Additionally, models in the simpler setup performed well and were trained with identical hyperparameters. Conversely, in the standard setup, while models were able to memorize the training samples and Q/A examples, the solutions learned by the pre-trained Pythia-70M and Flan-T5 Small does not generalize well to the validation and test Q/A, unlike the OPT model.


\subsection{Behavior Analysis}
Our analyses thus far have focused on prompting the model with a question and allowing it to generate an answer. Now, we examine how the models respond when prompted with a question followed by part of the answer. We experiment with the following combinations:
\begin{enumerate}[label=\Alph*.]
    \item The second document alone, skipping the first.
    \item The first document followed by the third, intentionally omitting the second.
    \item The first half of the first document only.
    \item The first half of the first document followed by the second document.
    \item The last document followed by the first.
\end{enumerate}
These scenarios were tested using OPT and Pythia models trained on our largest dataset. We find that in all cases except the last, the models continued the answer seamlessly. Specifically:
\begin{enumerate}[label=\Alph*.]
    \item They follow the second document with the third, then the fourth, etc.
    \item They follow the third document with the fourth, then the fifth, and so on.
    \item They follow the first half of the first document with the second half, then proceed to the second document, the third, and so forth.
    \item They follow the second document with the third, then the fourth, and so on.
    \item They follow the first document with the second, third, etc., but eventually skip some documents, including the last one.
\end{enumerate}
These results indicate that the position of tokens in the sequence is not a significant factor. Instead, the models demonstrate a robust ability to continue the sequence as long as the tokens are in a logical order.

\subsection{Comprehensive Analysis}
Reflecting on our experimental observations, we can gain insights into the capabilities and failures of these models. We've observed that, given sufficient scale, the documents recalled by the models are typically of the correct length (\autoref{fig:hist_num_sen}) and error-free (\autoref{fig:doc_len_x_accuracy}). Additionally, models trained under the simplified setup successfully recall information from a single training document (\autoref{fig:unseg_vs_seg}). Therefore, the issue appears not to lie in the content of the recalled documents but rather in the quantity of documents being recalled. Indeed, with improper scale, models seem incapable of recalling the correct number of documents, instead recalling a random number of documents (\autoref{fig:hist_num_docs}).

Interestingly, the smallest Pythia model performs better if fine-tuned starting from random weights rather than the pre-trained weights (\autoref{tab:scratch_vs_pretrained}), suggesting that the poor performance of the pre-trained weights cannot be completely attributed to an architectural reason. Instead, the issue partly appears to be with the pre-training weights failing to learn a solution that generalizes to the problem of recalling the correct number of documents, rather than merely memorizing the training samples. Why this discrepancy occurs, particularly in contrast to the larger pre-trained Pythia models remains unclear and warrants further research. Different hyperparameters could potentially enable the smaller models to generalize well to our problem, but it is uncertain if this can be achieved without severely degrading the language modeling capabilities of the pre-trained model.

Regarding Flan-T5, given that the smallest model fine-tuned from scratch performs as poorly as the one fine-tuned from pre-trained weights, the root cause of the poor performance could be either architectural or due to improper hyperparameters. Additionally, the size of the model appears to influence its performance. Since Flan-T5 follows an encoder-decoder architecture, unlike the decoder-only structures of models such as OPT and Pythia, its parameters are divided roughly equally between the encoder and decoder. Consequently, the second largest Flan-T5 model's decoder is comparable in size to that of the third smallest Pythia model, which coincides with the point where performance begins to improve for Pythia (as seen in \autoref{fig:unseg_vs_seg}). Models within the Pythia suite smaller than this threshold do not show significant performance gains. However, the smallest Pythia model, when trained from scratch, outperforms Flan-T5 under similar conditions. This highlights that architectural factors can hinder the emergence of capabilities for same sized models. As for scale, our hypothesis is that the smaller models lack the capacity to develop the necessary circuitry to perform this task, but further research will be necessary to pinpoint the exact cause and clarify the challenges faced by these smaller models.

\section{Discussion} \label{sec:discussion}



In addition to the questions raised in the previous section, the following additional questions should be considered.

\subsection{Language Modeling}
One aspect not addressed in this study is whether models can perform the given task while retaining their language modeling capabilities. Due to the size of the models examined, repetitive fine-tuning on the training documents is necessary for them to memorize the data, which leads to overfitting on the task. Ideally, experiments would need to be conducted on much larger models, incorporating the training documents into the pre-training corpus, followed by standard instruction tuning. One of the tasks in this tuning would involve recalling all documents related to a given topic. This approach would help determine if a model can accomplish this in a manner that is useful for solving problems. Unfortunately, we currently lack the computational resources to conduct such experiments, and hence we leave this for future work.

\subsection{Information Distribution}
We observed a notable performance gap between the standard and simplified setups, supporting the findings by  ~\citet{prato-etal-2023-epik}. Their research indicates that LLMs more easily recall multiple pieces of information when this information is contained in a single training sample rather than dispersed across multiple samples. This raises questions about how the distribution of topic-related information across multiple training documents affects an LLM's ability to gauge its knowledge. Particularly, the impact of this distribution on the internal mechanisms of the LLM is not well understood.

Numerous studies have shown that language models can memorize entire passages and documents within their weights, enabling them to recall this information during inference~\cite{2020arXiv201207805C,2022arXiv220207646C,NEURIPS2022_fa0509f4,NEURIPS2023_59404fb8,DEWYNTER2023100024,2024arXiv240511577C}. Consequently, the strong performance of models in the simpler setup, where they only need to recall information from a single document per topic, is not surprising.

However, it remains unclear why recalling information from multiple documents presents a greater challenge. Specifically, how is this information encoded within the model parameters~\cite{singh-etal-2020-bertnesia,dai-etal-2022-knowledge,2022arXiv220205262M} and how does dispersed information affect the recalling process? Understanding these mechanisms is crucial for improving the performance of language models, as many real-world problems necessitate recalling information from multiple training documents.

\subsection{Knowledge Awareness \& Understanding}
While we have demonstrated that some LLMs possess an awareness of the extent of their knowledge concerning the topics in our benchmark, this does not necessarily mean that these models can gauge their knowledge across any topic.

Determining whether LLMs can accurately assess the scope of their understanding of topics from their pre-training corpus requires further investigation. Topics in practice could cover a wide array of subjects, including individuals, locations, events, and concepts. However, we believe that the specific type of topic is likely not an influential factor.

The critical element, in our view, is the breadth of these topics, which relates to the amount of information relevant for each. Our findings did not show a decline in model performance when recalling up to eight documents. Nevertheless, this observation might change if the number of documents were significantly increased. Further research is necessary to explore the limits and capabilities of models in handling broader topics.

A more profound question is the extent to which LLMs understand the scope of their entire knowledge base, or at least subsets of it. Given the vast amount of information LLMs learn during training, comprehending the scope of this knowledge or its subsets seems incredibly challenging. Yet, it would be beneficial for a model to understand the extent of its own expertise.

Finally, it is important to note that understanding the scope of one’s knowledge concerning a topic does not imply an understanding of that topic itself. Whether LLMs truly comprehend the knowledge they have memorized is a different research question from ours and is an active area of investigation~\cite{Bender2021OnTD,li2022emergent,2023arXiv231002207G}.

\section{Conclusion} \label{sec:conclusion}
This study focused on determining whether LLMs possess an understanding of the span of their own knowledge on specific topics. Notably, we observed that all models, if scaled sufficiently, know how many documents are authored by the same person. Consequently, these LLMs know how much they know about these individuals; otherwise, they would sporadically recall too few or too many documents.

More specifically, we find that this capability emerges based on the model's architecture, its size, the dataset's size used for training, and the effectiveness of the pre-trained weights in learning a solution that generalizes, rather than simply memorizing the training samples.

To the best of our knowledge, this is the first paper to explore this capability in LLMs, demonstrating that certain models can assess the extent of their knowledge on specific topics. Further research is required to determine whether this phenomenon is common across LLMs.

Overall, our research contributes to a deeper understanding of the capabilities and inner workings of these models. Grasping how aware LLMs are of their own knowledge and identifying any limitations in this regard is crucial, as this feature enhances the usefulness and trustworthiness of intelligent systems. Additional research is necessary to continue exploring this aspect.

\section{Limitations}\label{sec:limitations}
The potential insights from testing larger open-source models could be valuable to the community. However, computational limitations prevent us from conducting these analyses. We hope to undertake such experiments in the future.

\section{Ethical Considerations}\label{sec:ethics}
This research utilizes large language models trained on extensive textual datasets. While such models have demonstrated exceptional ability in generation, it is critical to highlight the ethical considerations that the data used for training these models inherently contains human biases. These, in turn, can manifest in the models' outputs. As such, it is essential when deploying such models, to critically evaluate their outputs, keeping in mind the likelihood of underlying bias.

\section*{Acknowledgements}

Jerry Huang is supported by a Natural Sciences and Engineering Research Council of Canada (NSERC) Canada Graduate Scholarship and Fonds de Recherche du Qu\'{e}bec Nature et technologies (FRQNT) training scholarship.

Sarath Chandar is supported by the Canada CIFAR AI Chairs program, the Canada Research Chair in Lifelong Machine Learning, and the NSERC Discovery Grant.

This research was enabled in part by compute resources provided by Mila (\href{https://mila.quebec/}{mila.quebec}) and the Digital Research Alliance of Canada (\href{https://alliancecan.ca/}{alliancecan.ca}).

\bibliography{acl, custom}

\appendix

\section{Dataset Details} \label{app:dataset_details}

\subsection{Size}
    \begin{table*}[ht]
    \centering
    \begin{tabular}{rrrrr}
        \toprule
        Diarists & Total Documents & Train Q/A & Val Q/A & Test Q/A \\
        \midrule
        1K & 4,482 & 894 & 50 & 50 \\
        2K & 9,012 & 1,804 & 100 & 100 \\
        4K & 17,895 & 3,577 & 199 & 199 \\
        8K & 36,000 & 7,200 & 400 & 400 \\
        16K & 72,000 & 14,400 & 800 & 800 \\
        32K & 144,000 & 28,800 & 1,600 & 1,600 \\
        64K & 288,000 & 57,600 & 3,200 & 3,200 \\
        \bottomrule
    \end{tabular}
    \caption{Specifications for each dataset used in our experiments.} \label{tab:dataset_details}
\end{table*}

The details of each dataset used in our experiments are provided in \autoref{tab:dataset_details}. The training set of each dataset consists of all of the documents (diary entries), as well as the train Q/A pairs. The validation and test sets solely consist of Q/A pairs. There are no overlaps between the Q/A pairs in the training, validation and test sets. As previously mentioned, for each author, we generate 1 to 8 diary entries, where each entry contains 1 to 8 sentences (excluding the title).

\subsection{Attributes}
The list of attributes sampled for each diary entry along with possible values are presented in \autoref{tab:attributes}. Sampling of both the attribute and its value is always performed randomly. A document cannot contain the same attribute more than once, irrespective of its value.
\begin{table*}[ht!]
    \centering
    \begin{tabular}{ll}
        \toprule
        Attribute & Possible Values \\
        \midrule
        \texttt{Location} & \texttt{[City, Countryside]} \\
        \texttt{Time} & \texttt{[Morning, Evening]} \\
        \texttt{Weather} & \texttt{[Sunny, Rain]} \\
        \texttt{Mood} & \texttt{[Happy, Sad]} \\
        \texttt{Restfulness} & \texttt{[Tired, Rested]} \\
        \texttt{Stress Level} & \texttt{[Stressed, Relaxed]} \\
        \texttt{Physical Activity} & \texttt{[Running, Weight Training]} \\
        \texttt{Meditated} & \texttt{[Yes, No]}\\
        \bottomrule
    \end{tabular}
    \caption{List of attributes used in diary entries, along with their possible values.}
    \label{tab:attributes}
\end{table*}


\section{Model Suite Differences} \label{app:model_dif}
The following outlines the architectural differences between the OPT, Pythia, and Flan-T5 suite of models, emphasizing their unique characteristics and training details.

Both OPT and Pythia are based on the GPT-3 architecture, with only slight variations. OPT employs learned positional embeddings and utilizes the ReLU activation function~\cite{relu}. In contrast, Pythia incorporates rotary positional embeddings and the GELU activation function~\cite{gelu}. A notable distinction in Pythia's architecture is its use of parallel residual connections, where the self-attention and feed-forward blocks run concurrently, and their outputs are summed along with the residual. This differs from OPT's sequential arrangement, where the self-attention block is followed by the feed-forward block. Additionally, Pythia forgoes the application of dropout after the attention and feed-forward blocks, unlike OPT, which applies a dropout rate of 0.1.

Turning to Flan-T5, this model remains largely faithful to the original Transformer architecture~\cite{2017arXiv170603762V}, with a few key exceptions. Layer normalization~\cite{2016arXiv160706450L} in Flan-T5 is applied before the residual, self-attention, and feed-forward blocks. In contrast, OPT and Pythia place the residual connections before the layer normalization. Flan-T5 also does not include a bias term in the layer normalization and adopts relative positional embeddings.

Regarding pre-training, OPT is trained on The Pile~\cite{ThePile} along with other datasets, whereas Pythia is exclusively trained on The Pile. Flan-T5 is a fine-tuned version of T5~\cite{T5}, with both models being trained on a mix of datasets. For more detailed information on the pre-training specifics and hyperparameters, readers are encouraged to refer to the respective papers for each model~\cite{opt,pythia,flan-t5}.

\section{Training Details} \label{app:training_details}
\begin{table*}[ht]
    \centering
    \begin{tabular}{lrr}
        \toprule
        Model & Parameters & LR \\
        \midrule
        OPT 7M & 7,490,560 & 4e-4 \\
        OPT 125M & 125,239,296 & 6e-5 \\
        OPT 350M & 331,196,416 & 3e-5 \\
        OPT 1.3B & 1,315,758,080 & 2e-5 \\
        OPT 2.7B & 2,651,596,800 & 1.6e-5 \\
        Flan-T5 Small & 76,961,152 & 1e-4 \\
        Flan-T5 Base & 247,577,856 & 1e-4 \\
        Flan-T5 Large & 783,150,080 & 1e-4 \\
        Flan-T5 XL & 2,849,757,184 & 1e-4 \\
        Pythia 70M & 70,426,624 & 1e-4 \\
        Pythia 160M & 162,322,944 & 6e-5 \\
        Pythia 410M & 405,334,016 & 3e-5 \\
        Pythia 1B & 1,011,781,632 & 3e-5 \\
        Pythia 1.4B & 1,414,647,808 & 2e-5 \\
        Pythia 2.8B & 2,775,208,960 & 1.6e-5 \\
        \bottomrule
    \end{tabular}
    \caption{Model size and learning rate used to fine-tune each model in our experiments.} \label{tab:model_hyperparameters}
\end{table*}

We train our models until they converge by employing the Adam optimizer~\cite{Adam}, which is configured with beta values of 0.9 and 0.999, and an epsilon of 1e-8. No weight decay is applied in this process. The learning rate is initially set to zero and then linearly increased to reach the model-specific rate detailed in \autoref{tab:model_hyperparameters} over the course of 3,600 steps. After this warm-up period, the learning rate is maintained constant. We set the batch size to 32.



\end{document}